\documentclass[11pt]{article}
\usepackage{acl2016}
\usepackage{times}
\usepackage{latexsym}
\usepackage{amsmath}
\usepackage{amssymb}
\usepackage{stmaryrd}
\usepackage{graphicx}
\usepackage{ifthen}
\usepackage{color}
\usepackage{xspace}
\usepackage[utf8]{inputenc}
\usepackage{multirow}
\usepackage[hyphens]{url}

\newcommand\sQ{\ensuremath{\mathcal{Q}}}

\newcommand\bI{\ensuremath{\mathbf{I}}}

\newcommand\bR{\ensuremath{\mathbf{R}}}

\newcommand\FigTop[4]{\begin{figure}[t] \begin{center} \includegraphics[scale=#2]{#1} \end{center} \caption{\label{fig:#3} #4} \end{figure}}

\newcommand\interpret[1]{\llbracket #1 \rrbracket}

\newcommand\refsec[1]{Section~\ref{sec:#1}}

\newcommand\reffig[1]{Figure~\ref{fig:#1}}

\newcommand\reftab[1]{Table~\ref{tab:#1}}

\newcommand\Section[2]{\section{#2}\label{sec:#1}}
\newcommand\Subsection[2]{\subsection{#2}\label{sec:#1}}

\ifthenelse{\isundefined{\definition}}{\newtheorem{definition}{Definition}}{}
\ifthenelse{\isundefined{\assumption}}{\newtheorem{assumption}{Assumption}}{}
\ifthenelse{\isundefined{\hypothesis}}{\newtheorem{hypothesis}{Hypothesis}}{}
\ifthenelse{\isundefined{\proposition}}{\newtheorem{proposition}{Proposition}}{}
\ifthenelse{\isundefined{\theorem}}{\newtheorem{theorem}{Theorem}}{}
\ifthenelse{\isundefined{\lemma}}{\newtheorem{lemma}{Lemma}}{}
\ifthenelse{\isundefined{\corollary}}{\newtheorem{corollary}{Corollary}}{}
\ifthenelse{\isundefined{\alg}}{\newtheorem{alg}{Algorithm}}{}
\ifthenelse{\isundefined{\example}}{\newtheorem{example}{Example}}{}

\newcommand\citep\cite
\newcommand\citet\newcite

\definecolor{gray}{gray}{0.35}
\definecolor{darkgreen}{rgb}{0.0, 0.5, 0.0}

\renewcommand\FigTop[4]{\begin{figure}[t]\centering\includegraphics[scale=#2]{#1}\vspace*{-.6em}\caption{\label{fig:#3} #4} \end{figure}}

\DeclareMathOperator*{\argmin}{arg\,min}

\newcommand\dataset{{\sc WikiTableQuestions}\xspace}

\newcommand\T[1]{\text{\texttt{#1}}}
\newcommand\C[1]{\text{\emph{#1}}}

\newcommand\G[1]{\textcolor{gray}{(#1)}}

\newcommand\nl[1]{``\emph{#1}''}

\newcommand\yay{92.1}

\aclfinalcopy
\def\aclpaperid{624}

\title{Inferring Logical Forms From Denotations}

\author{
  Panupong Pasupat \\
  Computer Science Department \\
  Stanford University \\
  {\tt ppasupat@cs.stanford.edu}
\And
	Percy Liang \\
  Computer Science Department \\
  Stanford University \\
  {\tt pliang@cs.stanford.edu}
}

\date{}

\begin{document}

\maketitle

\renewcommand*\ttdefault{cmtt}

\begin{abstract}
A core problem in learning semantic parsers from denotations is
picking out
consistent logical forms---those that yield the correct
denotation---from a combinatorially large space.
To control the search space,
previous work relied on restricted set of rules, which limits expressivity.
In this paper,
we consider a much more expressive class of logical forms,
and show how to use dynamic programming to
efficiently represent the complete set of consistent logical forms.
Expressivity also introduces many more
spurious logical forms
which are consistent with the correct
denotation but do not
represent the meaning of the utterance.
To address this, we generate fictitious worlds and 
use crowdsourced
denotations
on these worlds
to filter out spurious logical forms.
On the \dataset dataset,
we increase the coverage of answerable questions
from 53.5\% to 76\%, 
and the additional crowdsourced supervision
lets us rule out \yay\% of spurious logical forms.

\end{abstract}

\Section{intro}{Introduction}

Consider the task of learning to answer complex
natural language questions
(e.g., \nl{Where did the last 1st place finish occur?})
using only question-answer pairs
as supervision
\citep{clarke10world,liang11dcs,berant2013freebase,artzi2013uw}.
Semantic parsers map the question
into a logical form
(e.g., $\bR[\T{Venue}].\T{argmax}(\T{Position}.\T{1st}, \T{Index})$)
that can be executed on a knowledge source to obtain the answer (denotation).
Logical forms are very expressive
since they can be recursively composed,
but
this very expressivity makes
it more difficult to 
search over the space of logical forms.
Previous work sidesteps this obstacle by restricting
the set of possible logical form compositions,
but this is limiting.
For instance, for the system in \newcite{pasupat2015compositional},
in only 53.5\% of the examples
was the correct logical form even in the
set of generated logical forms.

The goal of this paper is to solve
two main challenges
that prevent us from generating more expressive logical forms.
The first challenge is computational:
the number of logical forms
grows exponentially as their size increases.
Directly enumerating over all logical forms
becomes infeasible,
and pruning techniques such as beam search
can inadvertently prune out correct logical forms.

The second challenge is the large increase in
\emph{spurious} logical forms---those that do not reflect the semantics of the question
but coincidentally execute to the correct denotation.
For example,
while logical forms $z_1,\dots,z_5$ in \reffig{running}
are all \emph{consistent} (they execute to the correct answer $y$),
the logical forms $z_4$ and $z_5$ are spurious
and would give incorrect answers if the table were to change.

\begin{figure}[tb!]\centering
{\small
\begin{tabular}{|c|c|c|c|c|} \hline
\textbf{Year} & \textbf{Venue} & \textbf{Position} & \textbf{Event} & \textbf{Time} \\ \hline
2001 & Hungary & 2nd & 400m & 47.12 \\
2003 & Finland & 1st & 400m & 46.69 \\
2005 & Germany & 11th & 400m & 46.62 \\
2007 & Thailand & 1st & relay & 182.05 \\
2008 & China & 7th & relay & 180.32 \\ \hline
\end{tabular}}\\[0.5em]
\begin{tabular}{r@{\; }l}
$x$: & \nl{Where did the last 1st place finish occur?} \\
$y$: & Thailand \\
\end{tabular}\\[0.5em]
{\small
\fbox{\parbox{7.5cm}{
\centering\textbf{Consistent}\\[0.4em]
\bgroup\def\arraystretch{1.2}
\begin{tabular}{|r@{\; }p{6.5cm}|}\hline
\multicolumn{2}{|c|}{\textbf{Correct}} \\
$z_1$: & {$\bR[\T{Venue}].\T{argmax}(\T{Position}.\T{1st}, \T{Index})$} \\
& {\hspace{0em}\color{gray} Among rows with Position = 1st, pick the one with maximum index, then return the Venue of that row.} \\
$z_2$: & {$\bR[\T{Venue}].\T{Index}.\T{max}(\bR[\T{Index}].\T{Position}.\T{1st})$} \\
& {\hspace{0em}\color{gray} Find the maximum index of rows with Position = 1st, then return the Venue of the row with that index.} \\
$z_3$: & {$\bR[\T{Venue}].\T{argmax}(\T{Position}.\T{Number}.\C{1},$} \\ &\hspace*{2.4cm} {$\bR[\lambda x.\bR[\T{Date}].\bR[\T{Year}].x])$} \\
& {\hspace{0em}\color{gray} Among rows with Position number 1, pick one with latest date in the Year column and return the Venue.} \\
\hline
\multicolumn{2}{|c|}{\textbf{Spurious}} \\
$z_4$: & {$\bR[\T{Venue}].\T{argmax}(\T{Position}.\T{Number}.\C{1},$} \\ &\hspace*{2.4cm} {$\bR[\lambda x.\bR[\T{Number}].\bR[\T{Time}].x])$} \\
& {\hspace{0em}\color{gray} Among rows with Position number 1, pick the one with maximum Time number. Return the Venue.} \\
$z_5$: & {$\bR[\T{Venue}].\T{Year}.\T{Number}.($} \\ &\hspace*{-.4cm} {$\bR[\T{Number}].\bR[\T{Year}].\T{argmax}(\T{Type}.\T{Row}, \T{Index}) - \C{1})$} \\
       & {\hspace{0em}\color{gray} Subtract 1 from the Year in the last row, then return the Venue of the row with that Year.} \\
\hline
\end{tabular}
\egroup
}}\vspace*{-0.15em}
\fbox{\parbox{7.5cm}{
\begin{tabular}{r@{\; }p{6.5cm}}
\multicolumn{2}{c}{\textbf{Inconsistent}} \\
$\tilde z$: & {$\bR[\T{Venue}].\T{argmin}(\T{Position}.\T{1st}, \T{Index})$} \\
& {\hspace{0em}\color{gray} Among rows with Position = 1st, pick the one with minimum index, then return the Venue. (= Finland)}
\end{tabular}
}}}
\caption{
Six logical forms
generated from the question $x$.
The first five are \emph{consistent}:
they execute to the correct answer $y$.
Of those, \emph{correct} logical forms $z_1$, $z_2$, and $z_3$
are different ways to represent the semantics of $x$,
while \emph{spurious} logical forms $z_4$ and $z_5$
get the right answer $y$ for the wrong reasons.
}
\label{fig:running}
\end{figure}

We address these two challenges
by solving two interconnected tasks.
The first task, which addresses the computational challenge,
is to enumerate the set $Z$ 
of all consistent logical forms
given a question $x$, a knowledge source $w$ (``world''),
and the target denotation $y$ (\refsec{dpd}).
Observing that
the space of possible denotations grows
much more slowly than the space of logical forms,
we perform \emph{dynamic programming on denotations} (DPD)
to make search feasible.
Our method is guaranteed to find all consistent logical forms
up to some bounded size.

Given the set $Z$ of consistent logical forms,
the second task
is to filter out spurious logical forms from $Z$ (\refsec{fictitious}).
Using the property that spurious logical forms ultimately give a wrong answer
when the data in the world $w$ changes,
we create \emph{fictitious worlds} to test the denotations
of the logical forms in $Z$.
We use crowdsourcing to annotate the correct denotations
on a subset of the generated worlds.
To reduce the amount of annotation needed,
we choose the subset
that maximizes the expected information gain.
The pruned set of logical forms
would provide a stronger supervision signal
for training a semantic parser.

We test our methods on the \dataset
dataset of complex questions on Wikipedia tables.
We define a simple, general set of deduction rules
(\refsec{deductionrules}),
and use DPD to confirm that the rules
generate a correct logical form
in 76\% of the examples,
up from the 53.5\% in \citet{pasupat2015compositional}.
Moreover, unlike beam search,
DPD is guaranteed to find all consistent logical forms up to a bounded size.
Finally, by using annotated data on fictitious worlds,
we are able to prune out \yay\% of the spurious logical forms.

\Section{setup}{Setup}
\FigTop{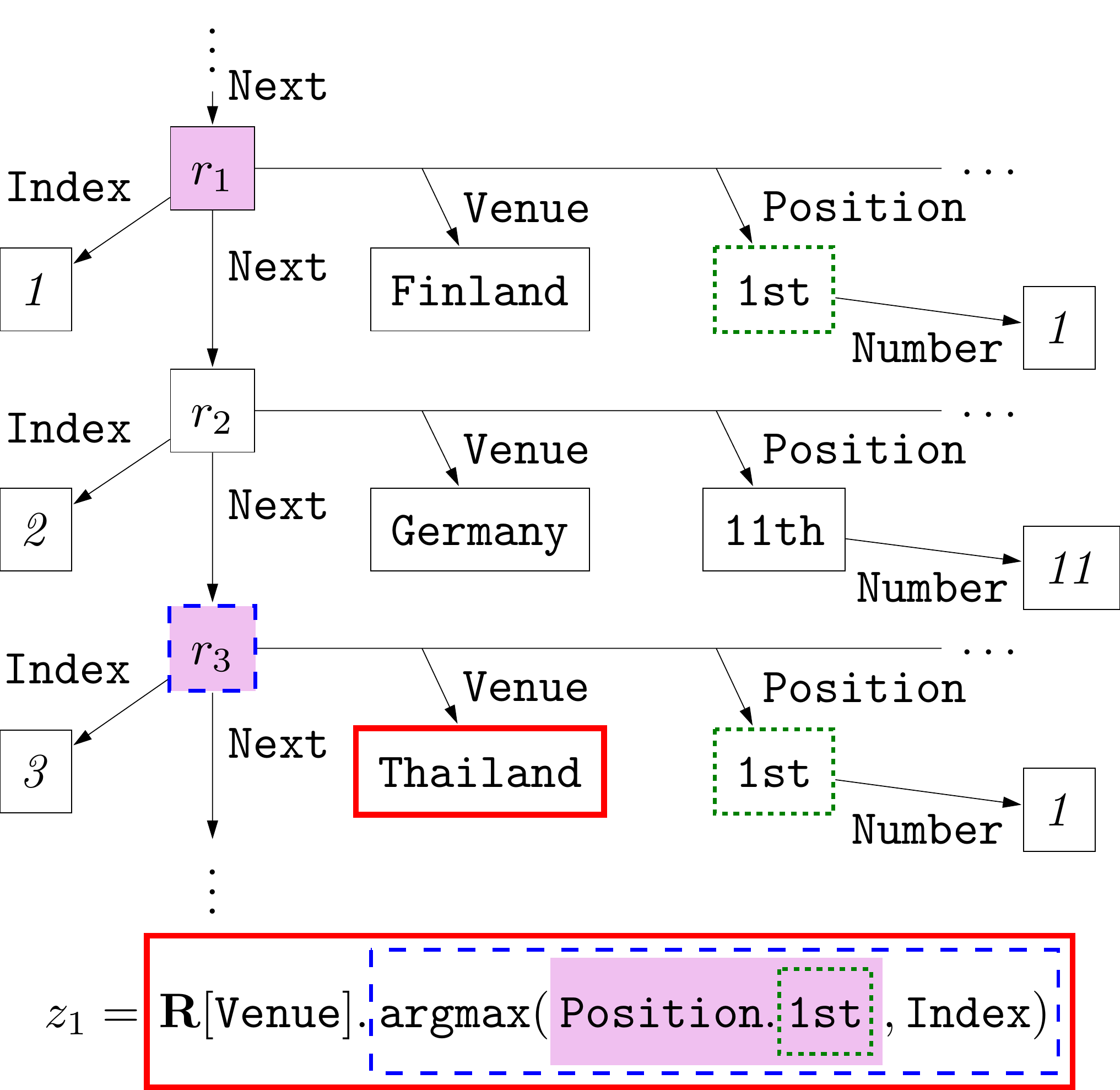}{0.35}{knowledgeGraph}
{
  The table in \reffig{running} is converted into a graph.
  The recursive execution of logical form $z_1$ is shown
  via the different colors and styles.
}

The overarching motivation of this work is allowing
people to ask questions involving computation
on semi-structured knowledge sources such as tables from the Web.
This section introduces how the knowledge source
is represented,
how the computation is carried out using \emph{logical forms},
and
our task of inferring correct logical forms.

\paragraph{Worlds.}
We use the term \emph{world} to refer to
a collection of
entities and relations between entities.
One way to represent a world $w$ is as
a directed graph
with nodes for entities and directed edges for relations.
(For example, a world about geography would contain
a node \T{Europe} with an edge \T{Contains} to another node \T{Germany}.)

In this paper, we use data tables from the Web
as knowledge sources,
such as the one in \reffig{running}.
We follow the construction in
\newcite{pasupat2015compositional}
for converting a table into a directed graph
(see \reffig{knowledgeGraph}).
Rows and cells become nodes (e.g., $r_0$ = first row and \T{Finland})
while columns become labeled directed edges between them
(e.g., \T{Venue} maps $r_1$ to \T{Finland}).
The graph is augmented with additional edges
\T{Next} (from each row to the next) and
\T{Index} (from each row to its index number).
In addition, we add normalization edges to cell nodes,
including 
\T{Number} (from the cell to the first number in the cell),
\T{Num2} (the second number),
\T{Date} (interpretation as a date),
and \T{Part} (each list item if the cell represents a list).
For example, a cell with content \nl{3-4}
has a \T{Number} edge to the integer \C{3}, a \T{Num2} edge to \C{4},
and a \T{Date} edge to \C{XX-03-04}.

\paragraph{Logical forms.}
We can perform computation on a world $w$ using
a \emph{logical form} $z$, a small program
that can be executed on the world,
resulting in a \emph{denotation} $\interpret{z}_w$.

We use lambda DCS \cite{liang2013lambdadcs}
as the language of logical forms.
As a demonstration,
we will use $z_1$ in \reffig{knowledgeGraph} as an example.
The smallest units of lambda DCS
are entities (e.g., \T{1st})
and relations (e.g., \T{Position}).
Larger logical forms
can be constructed using logical operations,
and the denotation of the new logical form can be
computed from denotations of its constituents.
For example, applying the \emph{join} operation
on \T{Position} and \T{1st} gives $\T{Position}.\T{1st}$,
whose denotation is the set of entities with relation \T{Position}
pointing to \T{1st}.
With the world in \reffig{knowledgeGraph}, the denotation is
$\interpret{\T{Position}.\T{1st}}_w = \{r_1, r_3\}$, which corresponds to the 2nd and 4th rows in the table.
The partial logical form $\T{Position}.\T{1st}$ is then used to construct
$\T{argmax}(\T{Position}.\T{1st}, \T{Index})$,
the denotation of which
can be computed by mapping the entities in $\interpret{\T{Position}.\T{1st}}_w = \{r_1, r_3\}$
using the relation $\T{Index}$ ($\{r_0: \C{0}, r_1: \C{1}, \dots\}$),
and then picking the one with the largest mapped value ($r_3$, which is mapped to \C{3}).
The resulting logical form is finally combined with $\bR[\T{Venue}]$ with another \emph{join} operation.
The relation $\bR[\T{Venue}]$ is the \emph{reverse} of $\T{Venue}$,
which corresponds to traversing $\T{Venue}$ edges in the reverse direction.

\paragraph{Semantic parsing.}
A semantic parser maps a natural language utterance $x$
(e.g., \nl{Where did the last 1st place finish occur?})
into a logical form $z$.
With denotations as supervision,
a semantic parser is trained to put high probability
on $z$'s that are \emph{consistent}---%
logical forms that execute to
the correct denotation $y$ (e.g., Thailand).
When the space of logical forms is large,
searching for consistent logical forms $z$ can become a challenge.

As illustrated in \reffig{running},
consistent logical forms can be divided into two groups:
\emph{correct} logical forms represent valid ways for computing the answer,
while \emph{spurious} logical forms accidentally get the right answer for the wrong reasons
(e.g., $z_4$ picks the row with the maximum time
but gets the correct answer anyway).

\paragraph{Tasks.}
Denote by $Z$ and $Z_c$ the sets of all consistent and
correct logical forms,
respectively.
The first task is to efficiently compute $Z$ given an utterance $x$,
a world $w$, and the correct denotation $y$ (\refsec{dpd}).
With the set $Z$,
the second task is to infer $Z_c$
by pruning spurious logical forms from $Z$ (\refsec{fictitious}).

\Section{deductionrules}{Deduction rules}

The space of logical forms
given an utterance $x$ and a world $w$
is defined recursively by a set of \emph{deduction rules}
(\reftab{ruletable}).
In this setting,
each constructed logical form belongs to a \emph{category}
(\C{Set}, \C{Rel}, or \C{Map}).
These categories are used for type checking
in a similar fashion to categories in syntactic parsing.
Each deduction rule specifies the categories of the arguments,
category of the resulting logical form,
and how the logical form is constructed from the arguments.

\newcommand\explain[1]{\multicolumn{4}{c}{\G{#1}}}

\begin{table}[tb]\centering\small
\begin{tabular}{@{\;}lr@{ $\to$ }ll@{}}
\multicolumn{3}{c}{\textbf{Rule}} & \textbf{Semantics} \\ \hline

\multicolumn{4}{c}{\textbf{\emph{Base Rules}}} \\

B1 &
$\C{TokenSpan}$ & $\C{Set}$ & $\mathrm{fuzzymatch}(span)$ \\
\explain{entity fuzzily matching the text: \nl{chinese} $\to$ \T{China}} \\

B2 &
$\C{TokenSpan}$ & $\C{Set}$ & $\mathrm{val}(span)$ \\
\explain{interpreted value: \nl{march 2015} $\to$ \C{2015-03-XX}} \\

B3 &
$\emptyset$ & $\C{Set}$ & $\T{Type}.\T{Row}$ \\
\explain{the set of all rows} \\

B4 &
$\emptyset$ & $\C{Set}$ & $c \in \mathrm{ClosedClass}$ \\
\explain{any entity from a column with few unique entities} \\
\explain{e.g., \T{400m} or \T{relay} from the Event column} \\

B5 &
$\emptyset$ & $\C{Rel}$ & $r \in \mathrm{GraphEdges}$ \\
\explain{any relation in the graph: \T{Venue}, \T{Next}, \T{Num2}, \dots} \\

B6 &
$\emptyset$ & $\C{Rel}$ & $\T{!=} \mid \T{<} \mid \T{<=} \mid \T{>} \mid \T{>=}$ \\

\hline

\multicolumn{4}{c}{\textbf{\emph{Compositional Rules}}} \\

C1 &
$\C{Set} + \C{Rel}$ & $\C{Set}$ & $z_2.z_1 \mid \bR[z_2].z_1$ \\ 
\explain{$\bR[z]$ is the reverse of $z$; i.e., flip the arrow direction} \\

C2 &
$\C{Set}$ & $\C{Set}$ & $a(z_1)$ \\
\explain{$a \in \{\T{count}, \T{max}, \T{min}, \T{sum}, \T{avg}\}$} \\

C3 &
$\C{Set} + \C{Set}$ & $\C{Set}$ & $z_1 \sqcap z_2 \mid z_1 \sqcup z_2 \mid z_1 - z_2$ \\
\explain{subtraction is only allowed on numbers}  \\

\hline

\multicolumn{4}{c}{\textbf{\emph{Compositional Rules with Maps}}} \\

\multicolumn{4}{c}{\textbf{Initialization}} \\

M1 &
$\C{Set}$ & $\C{Map}$ & $(z_1, x)$ \quad{\color{gray} (identity map)} \\

\multicolumn{4}{c}{\textbf{Operations on Map}} \\

M2 &
$\C{Map} + \C{Rel}$ & $\C{Map}$ & $(u_1, z_2.b_1) \mid (u_1, \bR[z_2].b_1)$ \\

M3 &
$\C{Map}$ & $\C{Map}$ & $(u_1, a(b_1))$ \\
\explain{$a \in \{\T{count}, \T{max}, \T{min}, \T{sum}, \T{avg}\}$} \\

M4 &
$\C{Map} + \C{Set}$ & $\C{Map}$ & $(u_1, b_1 \sqcap z_2) \mid \dots$ \\
M5 &
$\C{Map} + \C{Map}$ & $\C{Map}$ & $(u_1, b_1 \sqcap b_2) \mid \dots$ \\
\explain{Allowed only when $u_1 = u_2$} \\
\explain{Rules M4 and M5 are repeated for $\sqcup$ and $-$} \\

\multicolumn{4}{c}{\textbf{Finalization}} \\

M6 &
$\C{Map}$ & $\C{Set}$ & $\T{argmin}(u_1, \bR[\lambda x.b_1])$ \\
& \multicolumn{2}{c}{} & $\quad\mid \T{argmax}(u_1, \bR[\lambda x.b_1])$ \\

\hline

\end{tabular}
\caption{
Deduction rules define the space of logical forms
by specifying how partial logical forms are constructed.
The logical form of the $i$-th argument is denoted by $z_i$
(or $(u_i, b_i)$ if the argument is a \C{Map}).
The set of final logical forms contains any logical form
with category \C{Set}.
}\label{tab:ruletable}
\end{table}

Deduction rules are divided into
base rules and compositional rules.
A base rule follows one of the following templates:
\begin{align}
\C{TokenSpan}[span] &\to c\,[f(span)] \tag{1}\label{eqn:b1} \\
\emptyset &\to c\,[f()] \tag{2}\label{eqn:b2}
\end{align}
A rule of Template \ref{eqn:b1} is triggered by a span of tokens from $x$
(e.g., to construct $z_1$ in \reffig{knowledgeGraph}
from $x$ in \reffig{running},
Rule B1 from \reftab{ruletable} constructs \T{1st} of category \C{Set} from the phrase \nl{1st}).
Meanwhile, a rule of Template \ref{eqn:b2}
generates a logical form without any trigger
(e.g., Rule B5 generates \T{Position} of category \C{Rel}
from the graph edge \T{Position} without a specific trigger in $x$).

Compositional rules then
construct larger logical forms from smaller ones:
\begin{align}
c_1\,[z_1] + c_2\,[z_2] &\to c\,[g(z_1,z_2)] \tag{3}\label{eqn:c1} \\
c_1\,[z_1] &\to c\,[g(z_1)] \tag{4}\label{eqn:c2}
\end{align}
A rule of Template \ref{eqn:c1} 
combines partial logical forms $z_1$ and $z_2$
of categories $c_1$ and $c_2$ into
$g(z_1, z_2)$ of category $c$
(e.g., Rule C1 uses \T{1st} of category \C{Set}
and \T{Position} of category \C{Rel}
to construct $\T{Position}.\T{1st}$ of category \C{Set}).
Template \ref{eqn:c2} works similarly.

Most rules
construct logical forms without requiring a trigger from the utterance $x$.
This is crucial for generating implicit relations
(e.g., generating \T{Year} from \nl{what's the venue in 2000?}
without a trigger \nl{year}),
and generating operations without a lexicon
(e.g., generating \T{argmax} from \nl{where's the longest competition}).
However, the downside is that the space of possible logical forms becomes very large.

\paragraph{The \C{Map} category.}
The technique in this paper
requires execution of partial logical forms.
This poses a challenge for \T{argmin} and \T{argmax} operations,
which take a set and a binary relation as arguments.
The binary could be a complex function (e.g., in $z_3$ from \reffig{running}).
While it is possible to build the binary independently from the set,
executing a complex binary is sometimes impossible
(e.g., the denotation of $\lambda x.\T{count}(x)$ is impossible to write
explicitly without knowledge of $x$).

We address this challenge with
the $\C{Map}$ category.
A \C{Map} is a pair $(u, b)$ of a finite set $u$ (unary)
and a binary relation $b$.
The denotation of $(u, b)$ is $(\interpret{u}_w, \interpret{b}'_w)$
where the binary $\interpret{b}'_w$ is $\interpret{b}_w$
with the domain restricted to the set $\interpret{u}_w$.
For example, consider the construction of
$\T{argmax}(\T{Position}.\T{1st}, \T{Index})$.
After constructing $\T{Position}.\T{1st}$ with denotation $\{r_1, r_3\}$,
Rule M1 initializes $(\T{Position}.\T{1st}, x)$
with denotation $(\{r_1, r_3\}, \{r_1: \{r_1\}, r_3: \{r_3\}\})$.
Rule M2 is then applied to generate
$(\T{Position}.\T{1st}, \bR[\T{Index}].x)$
with denotation $(\{r_1, r_3\}, \{r_1: \{1\}, r_3: \{3\}\})$.
Finally, Rule M6 converts the \C{Map} into the desired \T{argmax} logical form
with denotation $\{r_3\}$.

\paragraph{Generality of deduction rules.}
Using domain knowledge,
previous work restricted the space of logical forms
by manually defining the categories $c$
or the semantic functions $f$ and $g$ to fit the domain.
For example,
the category $\C{Set}$ might be divided into
$\C{Records}$, $\C{Values}$, and $\C{Atomic}$
when the knowledge source is a table \cite{pasupat2015compositional}.
Another example
is when a compositional rule $g$ (e.g., $\T{sum}(z_1)$)
must be triggered by some phrase in a lexicon
(e.g., words like \nl{total}
that align to \T{sum} in the training data).
Such restrictions make search more tractable but
greatly limit the scope of questions that can be answered.

Here, we have increased the coverage of logical forms
by making the deduction rules simple and general,
essentially following the syntax of lambda DCS.
The base rules only generates entities that approximately match the utterance,
but all possible relations, and all possible further combinations.

\paragraph{Beam search.}

Given the deduction rules,
an utterance $x$ and a world $w$,
we would like to generate all derived logical forms $Z$.
We first present the floating parser \cite{pasupat2015compositional},
which uses beam search to generate $Z_\text{b} \subseteq Z$,
a usually incomplete subset.
Intuitively, the algorithm first constructs base logical forms
based on spans of the utterance,
and then builds larger logical forms of increasing size
in a ``floating'' fashion---without requiring a trigger from the utterance.

Formally, partial logical forms
with category $c$ and size $s$ are stored in a \emph{cell} $(c,s)$.
The algorithm first generates base logical forms
from base deduction rules
and store them in cells $(c, 0)$
(e.g., the cell $(\C{Set}, 0)$ contains \T{1st}, $\T{Type}.\T{Row}$, and so on).
Then for each size $s = 1,\dots,s_\mathrm{max}$,
we populate the cells $(c,s)$ by applying compositional rules
on partial logical forms with size less than $s$.
For instance, when $s = 2$, we can apply Rule C1 on
logical forms $\T{Number}.\C{1}$ from cell $(\C{Set},s_1 = 1)$
and $\T{Position}$ from cell $(\C{Rel},s_2 = 0)$
to create $\T{Position}.\T{Number}.\C{1}$
in cell $(\C{Set}, s_0 + s_1 + 1 = 2)$.
After populating each cell $(c,s)$,
the list of logical forms in the cell
is pruned
based on the model scores
to a fixed beam size
in order to control the search space.
Finally, the set $Z_\text{b}$
is formed by
collecting logical forms from all cells $(\C{Set},s)$
for $s = 1,\dots,s_{\rm max}$.

Due to the generality of our deduction rules,
the number of logical forms grows quickly as the size $s$ increases.
As such, partial logical forms that are essential
for building the desired logical forms
might fall off the beam early on.
In the next section, we present a new search method
that compresses the search space using denotations.

\Section{dpd}{Dynamic programming on denotations}

\FigTop{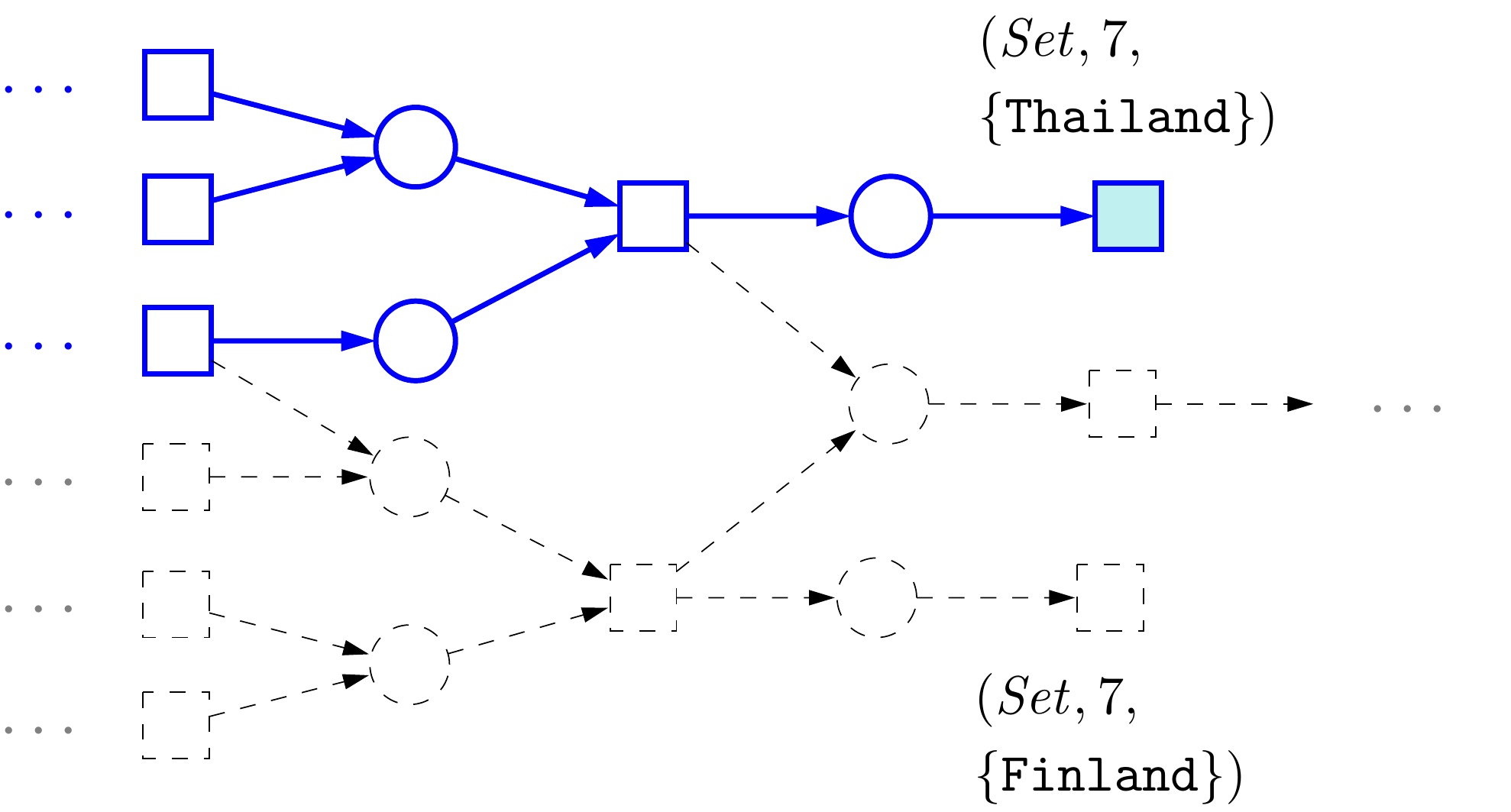}{0.35}{dpdConcept}
{
The first pass of DPD constructs cells $(c, s, d)$ (square nodes)
using denotationally invariant semantic functions (circle nodes).
The second pass enumerates all logical forms along paths that lead to the
correct denotation $y$ ({\color{blue}{solid lines}}).
}

Our first step toward finding all correct logical forms 
is to represent all consistent logical forms (those that execute to the correct denotation).
Formally, given $x$, $w$, and $y$, we wish to generate the set $Z$
of all logical forms $z$ such that $\interpret{z}_w = y$.

As mentioned in the previous section,
beam search does not recover the full set $Z$ due to pruning.
Our key observation is that 
while the number of logical forms explodes,
the number of \emph{distinct denotations}
of those logical forms is much more controlled,
as multiple logical forms can
share the same denotation.
So instead of directly enumerating logical forms,
we use \emph{dynamic programming on denotations} (DPD),
which is inspired by similar methods
from program induction \citep{lau03programming,liang10programs,gulwani2011automating}.

The main idea of DPD is to
collapse logical forms with the same denotation together.
Instead of using cells $(c, s)$ as in beam search,
we perform dynamic programming
using cells $(c, s, d)$ where $d$ is a denotation.
For instance, 
the logical form
$\T{Position}.\T{Number}.\C{1}$
will now be stored in cell $(\C{Set}, 2, \{r_1, r_3\})$.

For DPD to work, each deduction rule must have
a \emph{denotationally invariant} semantic function $g$,
meaning that the denotation of the resulting logical form $g(z_1,z_2)$
only depends on the denotations of $z_1$ and $z_2$:
\begin{multline*}
\interpret{z_1}_w = \interpret{z_1'}_w \;\wedge\;
\interpret{z_2}_w = \interpret{z_2'}_w \\
\qquad \Rightarrow \interpret{g(z_1,z_2)}_w = \interpret{g(z_1',z_2')}_w
\end{multline*}
All of our deduction rules in \reftab{ruletable} are denotationally invariant,
but a rule that, for instance, returns the argument with the larger logical form size would not be.
Applying a denotationally invariant deduction rule on any pair of logical forms
from $(c_1,s_1,d_1)$ and $(c_2,s_2,d_2)$ always results
in a logical form with the same denotation $d$
in the same cell $(c,s_1+s_2+1,d)$.%
\footnote{Semantic functions $f$ with one argument work similarly.}
(For example, the cell $(\C{Set}, 4, \{r_3\})$ contains
$z_1 := \T{argmax}(\T{Position}.\T{1st},\T{Index})$ and
$z_1' := \T{argmin}(\T{Event}.\T{Relay},\T{Index})$.
Combining each of these with $\T{Venue}$ using Rule C1 gives
$\bR[\T{Venue}].z_1$ and $\bR[\T{Venue}].z_1'$, which
belong to the same cell $(\C{Set}, 5, \{\T{Thailand}\})$).

\paragraph{Algorithm.}
DPD proceeds in two forward passes.
The first pass finds the possible combinations of cells $(c, s, d)$
that lead to the correct denotation $y$,
while the second pass enumerates the logical forms
in the cells found in the first pass.
\reffig{dpdConcept} illustrates the DPD algorithm.

In the first pass,
we are only concerned about finding relevant cell combinations
and not the actual logical forms.
Therefore, any logical form that belongs to a cell could be used
as an argument of a deduction rule to generate further logical forms.
Thus, we keep at most one logical form per cell;
subsequent logical forms that are generated for that cell are discarded.

After populating all cells up to size $s_\mathrm{max}$,
we list all cells $(\C{Set}, s, y)$
with the correct denotation $y$, and then
note all possible rule combinations $(\mathrm{cell}_1, \mathrm{rule})$
or $(\mathrm{cell}_1, \mathrm{cell}_2, \mathrm{rule})$
that lead to those final cells,
including the combinations that yielded discarded logical forms.

The second pass retrieves the actual logical forms
that yield the correct denotation.
To do this, we simply populate the cells $(c, s, d)$ with all logical forms,
using only rule combinations that lead to final cells.
This elimination of irrelevant rule combinations
effectively reduces the search space.
(In \refsec{ex-dpd}, we empirically show that
the number of cells considered is reduced by 98.7\%.)

The parsing chart is represented as a hypergraph
as in \reffig{dpdConcept}.
After eliminating unused rule combinations,
each of the remaining hyperpaths from base predicates
to the target denotation
corresponds to a single logical form.
making the remaining parsing chart a compact implicit representation
of all consistent logical forms.
This representation
is guaranteed to cover
all possible logical forms under the size limit $s_\mathrm{max}$
that can be constructed by the deduction rules.

In our experiments, we apply DPD on the deduction rules in \reftab{ruletable}
and
explicitly enumerate the logical forms produced by the second pass.
For efficiency, we prune logical forms that are clearly redundant
(e.g., applying \T{max} on a set of size 1).
We also restrict a few rules that might otherwise create too many denotations.
For example, we restricted
the union operation ($\sqcup$) except unions of two entities
(e.g., we allow $\T{Germany} \sqcup \T{Finland}$ but not $\T{Venue}.\T{Hungary} \sqcup \dots$),
subtraction when building a \C{Map},
and \T{count} on a set of size 1.%
\footnote{While we technically can apply \T{count} on sets of size 1,
the number of spurious logical forms explodes as there are
too many sets of size 1 generated.
}

\Section{fictitious}{Fictitious worlds}

After finding
the set $Z$ of all consistent logical forms,
we want to filter out spurious logical forms.
To do so, we observe that
semantically correct logical forms
should also give the correct denotation in worlds $w'$ other than than $w$.
In contrast, spurious logical forms will fail
to produce the correct denotation on some other world.

\paragraph{Generating fictitious worlds.}
With the observation above,
we generate \emph{fictitious worlds} $w_1, w_2, \dots$,
where each world $w_i$ is a slight alteration of $w$.
As we will be executing logical forms $z \in Z$ on $w_i$,
we should ensure that all 
entities and relations in $z \in Z$
appear in the fictitious world $w_i$
(e.g., $z_1$ in \reffig{running} would be meaningless
if the entity \T{1st} does not appear in $w_i$).
To this end, we impose that all
predicates present in the original world $w$
should also be present in $w_i$ as well.

In our case
where the world $w$ comes from a data table $t$,
we construct $w_i$ from a new table $t_i$ as follows:
we go through each column of $t$ and resample the cells in that column.
The cells are sampled using random draws without replacement
if the original cells are all distinct,
and with replacement otherwise.
Sorted columns are kept sorted.
To ensure that predicates in $w$ exist in $w_i$,
we use the same set of table columns
and enforce that any entity fuzzily matching a span in the question $x$
must be present in $t_i$
(e.g., for the example in \reffig{running},
the generated $t_i$ must contain ``1st'').
\reffig{fictitious1} shows an example fictitious table generated from the table
in \reffig{running}.

Fictitious worlds are similar to test suites for computer programs.
However, unlike manually designed test suites,
we do not yet know the correct answer for each fictitious world
or whether a world is helpful for filtering out spurious logical forms.
The next subsections introduce our method for choosing
a subset of useful fictitious worlds to be annotated.

\begin{figure}[tb!]\centering
{\small
\begin{tabular}{|c|c|c|c|c|} \hline
\textbf{Year} & \textbf{Venue} & \textbf{Position} & \textbf{Event} & \textbf{Time} \\ \hline
2001 & Finland & 7th & relay & 46.62 \\
2003 & Germany & 1st & 400m & 180.32 \\
2005 & China & 1st & relay & 47.12 \\
2007 & Hungary & 7th & relay & 182.05 \\ \hline
\end{tabular}}
\caption{From the example in \reffig{running},
we generate a table for the fictitious world $w_1$.
}\label{fig:fictitious1}
\end{figure}
\begin{figure}[tb!]\centering
\newcommand{\THA}{{\small Thailand}}
\newcommand{\CHN}{{\small China}}
\newcommand{\FIN}{{\small Finland}}
\newcommand{\GER}{{\small Germany}}
{
\begin{tabular}{c|c|c|c|c@{}c@{ }l}
\multicolumn{1}{c}{} & \multicolumn{1}{c}{$w$} &
\multicolumn{1}{c}{$w_1$} & \multicolumn{1}{c}{$w_2$} & $\cdots$ \\ \cline{2-4}
$z_1$ & \THA & \CHN & \FIN & $\cdots$ & \multirow{3}{*}{\Bigg\}} & \multirow{3}{*}{$q_1$} \\
$z_2$ & \THA & \CHN & \FIN & $\cdots$ & \\ 
$z_3$ & \THA & \CHN & \FIN & $\cdots$ & \\
$z_4$ & \THA & \GER & \CHN & $\cdots$ & \multirow{1}{*}{\}} & \multirow{1}{*}{$q_2$} \\
$z_5$ & \THA & \CHN & \CHN & $\cdots$ & \multirow{2}{*}{\Big\}} & \multirow{2}{*}{$q_3$} \\
$z_6$ & \THA & \CHN & \CHN & $\cdots$ & \\
$\vdots$ & $\vdots$ & $\vdots$ & $\vdots$ & \\ \cline{2-4}
\end{tabular}}
\caption{We execute consistent logical forms $z_i \in Z$
on fictitious worlds to get denotation tuples.
Logical forms with the same denotation tuple
are grouped into the same equivalence class $q_j$.}
\label{fig:equivClassTable}
\end{figure}

\paragraph{Equivalence classes.}
Let $W = (w_1, \dots, w_k)$ be the list of all possible fictitious worlds.
For each $z \in Z$, we define the denotation tuple
$\interpret{z}_W = (\interpret{z}_{w_1},\dots,\interpret{z}_{w_k})$.
We observe that some logical forms produce the same denotation
across all fictitious worlds.
This may be due to an algebraic
equivalence in logical forms
(e.g., $z_1$ and $z_2$ in \reffig{running})
or due to the constraints in the construction of fictitious worlds
(e.g., $z_1$ and $z_3$ in \reffig{running} are equivalent
as long as the Year column is sorted).
We group logical forms into equivalence classes
based on their denotation tuples,
as illustrated in \reffig{equivClassTable}.
When the question is unambiguous,
we expect at most one equivalence class
to contain correct logical forms.

\paragraph{Annotation.}
To pin down the correct equivalence class,
we acquire the correct answers to the question $x$ on some subset
$W' = (w'_1, \dots, w'_\ell) \subseteq W$ of $\ell$ fictitious worlds,
as it is impractical to obtain annotations on all fictitious worlds in $W$.
We compile equivalence classes that agree with the annotations
into a set $Z_\text{c}$ of correct logical forms.

We want to choose $W'$ that gives us the most information
about the correct equivalence class as possible.
This is analogous to standard practices in active learning \citep{settles2010active}.\footnote{
The difference is that we are obtaining partial information about an individual example
rather than partial information about the parameters.}
Let $\sQ$ be the set of all equivalence classes $q$,
and let $\interpret{q}_{W'}$ be the denotation tuple
computed by executing an arbitrary $z\in q$ on $W'$.
The subset $W'$ divides $\sQ$ into partitions
$F_t = \{q\in\sQ:\interpret{q}_{W'} = t\}$
based on the denotation tuples $t$
(e.g., from \reffig{equivClassTable},
if $W'$ contains just $w_2$, then $q_2$ and $q_3$ will be in the same partition $F_{(\text{China})}$).
The annotation $t^*$, which is also a denotation tuple, will mark
one of these partitions $F_{t^*}$ as correct.
Thus, to prune out many spurious equivalence classes,
the partitions should be as numerous and as small as possible.

More formally,
we choose a subset $W'$ that maximizes the expected information gain
(or equivalently, the reduction in entropy) about the correct equivalence class
given the annotation.
With random variables $Q \in \sQ$
representing the correct equivalence class and $T^*_{W'}$ for the annotation on worlds $W'$,
we seek to find
$\argmin_{W'} \operatorname{H}(Q\mid T^*_{W'})$.
Assuming a uniform prior on $Q$ ($p(q) = 1/|\sQ|$)
and accurate annotation ($p(t^*\mid q) = \bI[q\in F_{t^*}]$):
\begin{align*}
\operatorname{H}(Q \mid T^*_{W'})
&= \sum_{q,t} p(q,t) \log \frac{p(t)}{p(q,t)} \\
&= \frac{1}{|\sQ|} \sum_t |F_t| \log |F_t|. \tag{*}\label{eqn:entropy}
\end{align*}

We exhaustively search for $W'$ that minimizes (\ref{eqn:entropy}).
The objective value follows our intuition
since $\sum_t |F_t| \log |F_t|$ is small when
the terms $|F_t|$ are small and numerous.

In our experiments,
we approximate the full set $W$ of fictitious worlds by generating
$k = 30$ worlds to compute equivalence classes.
We choose a subset of $\ell = 5$ worlds to be annotated.

\Section{experiments}{Experiments}
\newcommand\previousWork{PL15\xspace}

For the experiments,
we use the training portion of the \dataset dataset
\cite{pasupat2015compositional},
which consists of 14,152 questions
on 1,679 Wikipedia tables gathered by crowd workers.
Answering these complex questions requires different types of operations.
The same operation can be phrased in different ways
(e.g., \nl{best}, \nl{top ranking}, or \nl{lowest ranking number})
and the interpretation of some phrases depend on the context 
(e.g., \nl{number of} could be a table lookup or a count operation).
The lexical content of the questions is also quite diverse:
even excluding numbers and symbols,
the 14,152 training examples contain 9,671 unique words,
only 10\% of which appear more than 10 times.

We attempted to manually annotate the first 300 examples
with lambda DCS logical forms.
We successfully constructed correct logical forms for 84\% of these examples,
which is a good number considering the questions were created
by humans who could use the table however they wanted.
The remaining 16\% reflect limitations in
our setup%
---for example,
non-canonical table layouts,
answers appearing in running text or images,
and common sense reasoning
(e.g., knowing that ``Quarter-final'' is better than ``Round of 16'').

\Subsection{ex-deductionrules}{Generality of deduction rules}
We compare our set of deduction rules with the
one given in \newcite{pasupat2015compositional} (henceforth \previousWork).
\previousWork{} reported generating
the annotated logical form in 53.5\% of the first 200 examples.
With our more general deduction rules,
we use DPD to verify that the rules are able to generate
the annotated logical form in
76\% of the first 300 examples,
within the logical form size limit $s_\mathrm{max}$ of 7.
This is 90.5\% of the examples that were successfully annotated.
\reffig{newLFs} shows some examples of logical forms we cover that \previousWork{} could not.
Since DPD is guaranteed to find all consistent logical forms,
we can be sure that the logical forms not covered
are due to limitations of the deduction rules.
Indeed, the remaining examples either have logical forms with size larger than 7
or require other operations such as addition, union of arbitrary sets, etc.

\begin{figure}[tb!]\centering
{\small
\bgroup\def\arraystretch{1.2}
\begin{tabular}{|l|} \hline
\nl{which opponent has the most wins} \\
$z = \T{argmax}(\bR[\T{Opponent}].\T{Type}.\T{Row},$ \\
\qquad \quad $\bR[\lambda x.\T{count}(\T{Opponent}.x \sqcap \T{Result}.\T{Lost}])$ \\
\hline
\nl{how long did ian armstrong serve?} \\
$z = \bR[\T{Num2}].\bR[\T{Term}].\T{Member}.\T{IanArmstrong}$ \\
\qquad \quad $-\,\bR[\T{Number}].\bR[\T{Term}].\T{Member}.\T{IanArmstrong}$ \\
\hline
\nl{which players came in a place before lukas bauer?} \\
$z = \bR[\T{Name}].\T{Index}.\T{<}.\bR[\T{Index}].\T{Name}.\T{LukasBauer}$ \\
\hline
\nl{which players played the same position as ardo kreek?} \\
$z = \bR[\T{Player}].\T{Position}.\bR[\T{Position}].\T{Player}.\T{Ardo}$ \\
\qquad \quad $ \sqcap\;\T{!=}.\T{Ardo}$ \\
\hline
\end{tabular}
\egroup
\vspace*{-0.6em}
}
\caption{Several example logical forms our system can generated
that are not covered by the deduction rules from the previous work
\previousWork.}
\label{fig:newLFs}
\end{figure}

\Subsection{ex-dpd}{Dynamic programming on denotations}
\paragraph{Search space.}
To demonstrate the savings gained by collapsing logical forms with the same denotation,
we track the growth of the number of unique logical forms and denotations
as the logical form size increases.
The plot in
\reffig{exGrowth} shows that
the space of logical forms explodes much more quickly
than the space of denotations.

The use of denotations also saves us from considering
a significant amount of irrelevant partial logical forms.
On average over 14,152 training examples,
DPD generates approximately
25,000
consistent logical forms.
The first pass of DPD generates $\approx$~%
153,000 cells $(c,s,d)$,
while the second pass
generates only $\approx$~%
2,000
cells
resulting from $\approx$~%
8,000
rule combinations,
resulting in a 98.7\% reduction in the number of cells
that have to be considered.

\paragraph{Comparison with beam search.}
We compare DPD to beam search
on the ability to generate (but not rank) the annotated logical forms.
We consider two settings:
when the beam search parameters
are uninitialized (i.e., the beams are pruned randomly),
and when the parameters are trained using the system from \previousWork
(i.e., the beams are pruned based on model scores).
The plot in
\reffig{exFloat} shows that
DPD generates more annotated logical forms (76\%)
compared to beam search (53.7\%),
even when beam search is guided heuristically by learned parameters.
Note that DPD is an exact algorithm and does not require a heuristic.

\FigTop{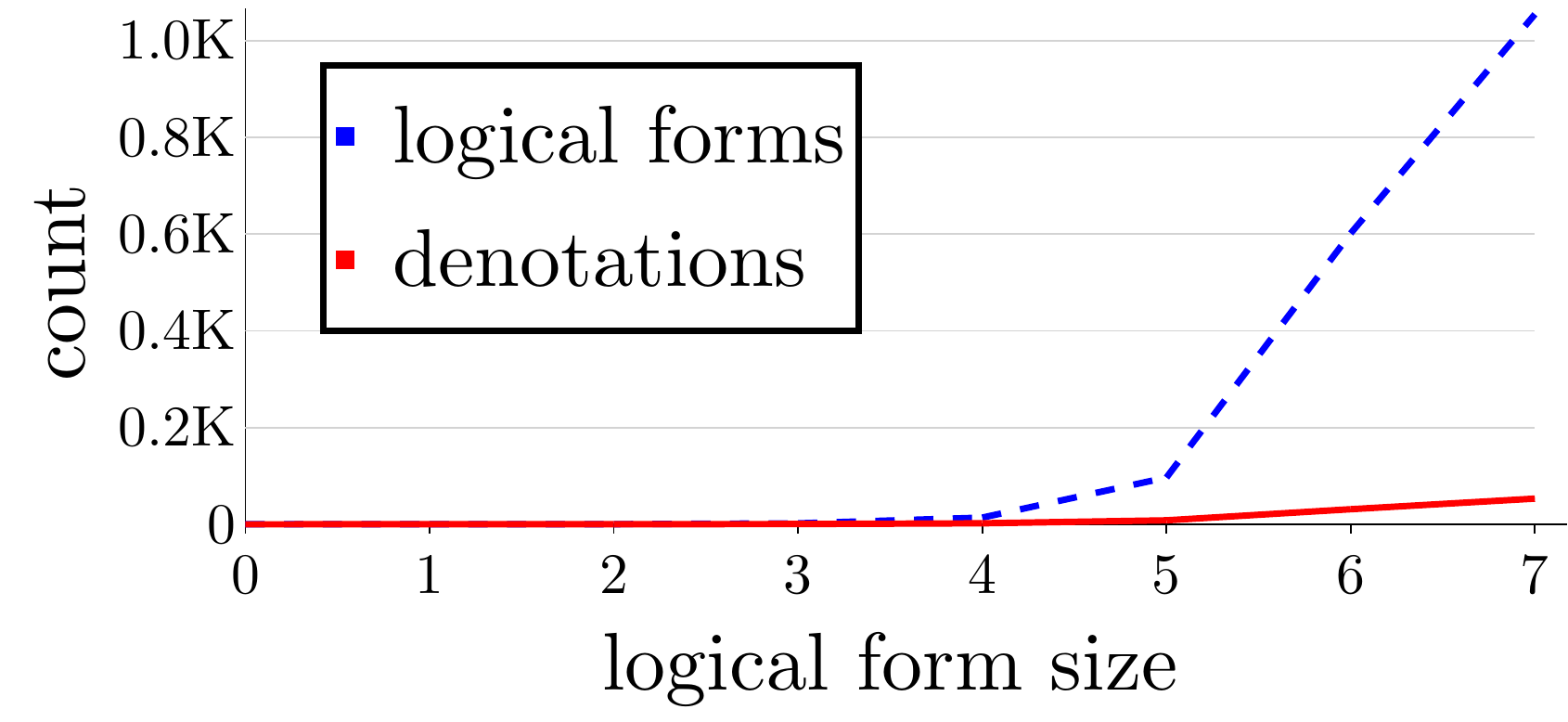}{0.35}{exGrowth}
{The median of the number of logical forms ({\color{blue}dashed}) and denotations ({\color{red}solid})
as the formula size increases.
The space of logical forms grows much faster than the space of denotations.}

\FigTop{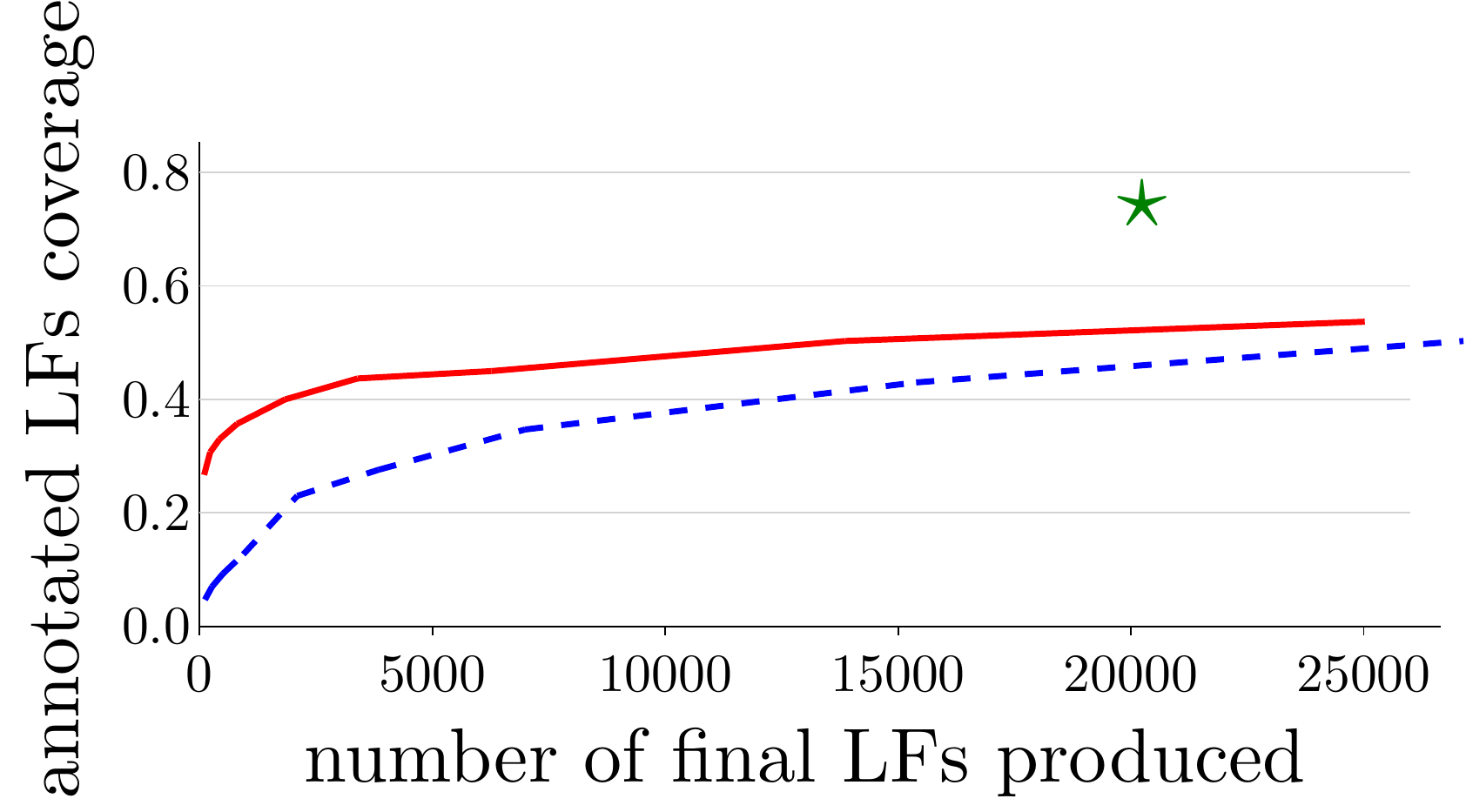}{0.32}{exFloat}
{The number of annotated logical forms that can be generated by
beam search, both uninitialized ({\color{blue}dashed}) and
initialized ({\color{red}solid}),
increases with the number of candidates generated
(controlled by beam size), but lacks behind
DPD ({\color{darkgreen}star}).}

\Subsection{ex-fictitious}{Fictitious worlds}
We now explore how fictitious worlds divide the set of logical forms
into equivalence classes,
and how the annotated denotations on the chosen worlds
help us prune spurious logical forms.

\paragraph{Equivalence classes.}
Using 30 fictitious worlds per example,
we produce an average of 1,237 equivalence classes.
One possible concern with using a limited number of fictitious worlds
is that we may fail to distinguish some pairs of
non-equivalent logical forms.
We verify the equivalence classes
against the ones computed using 300 fictitious worlds.
We found that only $5\%$ of the logical forms
are split from the original equivalence classes.

\paragraph{Ideal Annotation.}
After computing equivalence classes,
we choose a subset $W'$ of 5 fictitious worlds to be annotated
based on the information-theoretic objective.
For each of the 252 examples with an annotated logical form $z^*$,
we use the denotation tuple $t^* = \interpret{z^*}_{W'}$ as the annotated answers on the chosen fictitious worlds.
We are able to rule out 98.7\% of the spurious equivalence classes
and 98.3\% of spurious logical forms.
Furthermore,
we are able to filter down to just one equivalence class in
32.7\% of the examples,
and at most three equivalence classes in
51.3\% of the examples.
If we choose 5 fictitious worlds randomly instead of maximizing information gain,
then the above statistics are
22.6\%
and
36.5\%,
respectively.
When more than one equivalence classes remain,
usually only one class is a dominant class
with many equivalent logical forms,
while other classes are small and contain
logical forms with unusual patterns
(e.g., $z_5$ in \reffig{running}).

The average size of the correct equivalence class is $\approx$~3,000
with the standard deviation of $\approx$~8,000.
Because we have an expressive logical language,
there are fundamentally
many equivalent ways of computing the same quantity.

\paragraph{Crowdsourced Annotation.}
Data from crowdsourcing is more susceptible to errors.
From the 252 annotated examples,
we use 177 examples where at least two crowd workers agree on the answer of the original world $w$.
When the crowdsourced data is used to rule out spurious logical forms,
the entire set $Z$ of consistent logical forms is pruned out in 11.3\% of the examples,
and the correct equivalent class is removed in 9\% of the examples.
These issues are due to annotation errors,
inconsistent data (e.g., having date of death before birth date),
and different interpretations of the question on the fictitious worlds.
For the remaining examples, we are able to prune out \yay\% of spurious logical forms
(or 92.6\% of spurious equivalence classes).

To prevent the entire $Z$ from being pruned,
we can relax our assumption and 
keep logical forms $z$ that disagree with the annotation in at most 1 fictitious world.
The number of times $Z$ is pruned out is reduced to 3\%,
but the number of spurious logical forms pruned also decreases to 78\%.

\Section{discussion}{Related Work and Discussion}

This work evolved from a long tradition of learning executable semantic parsers,
initially from annotated logical forms
\cite{zelle96geoquery,kate05funql,zettlemoyer05ccg,zettlemoyer07relaxed,kwiatkowski10ccg},
but more recently from denotations
\cite{clarke10world,liang11dcs,berant2013freebase,kwiatkowski2013scaling,pasupat2015compositional}.
A central challenge in learning from denotations is finding consistent logical forms
(those that execute to a given denotation).

As \citet{kwiatkowski2013scaling} and \citet{berant2014paraphrasing} both noted,
a chief difficulty with executable semantic parsing
is the ``schema mismatch''---words in the utterance do not map cleanly onto the predicates
in the logical form.
This mismatch is especially pronounced in the \dataset of \citet{pasupat2015compositional}.
In the second example of \reffig{newLFs}, \nl{how long} is realized by a logical form
that computes a difference between two dates.
The ramification of this mismatch is that finding consistent logical forms
cannot solely proceed from the language side.
This paper is about using annotated denotations to drive the search over logical forms.

This takes us into the realm of program induction,
where the goal is to infer a program (logical form) from input-output pairs
(for us, world-denotation pairs).
Here, previous work has also leveraged the idea of dynamic programming on denotations
\citep{lau03programming,liang10programs,gulwani2011automating},
though for more constrained spaces of programs.
Continuing the program analogy,
generating fictitious worlds is similar in spirit to fuzz testing for
generating new test cases \citep{miller1990empirical},
but the goal there is coverage in a single program rather than
identifying the correct (equivalence class of) programs.
This connection can potentially improve the flow of ideas between the two fields.

Finally, the effectiveness of dynamic programming on denotations
relies on having a manageable set of denotations.
For more complex logical forms and larger knowledge graphs,
there are many possible angles worth exploring:
performing abstract interpretation to collapse denotations into equivalence
classes \citep{cousot77abstract},
relaxing the notion of getting the correct denotation \citep{steinhardt2015relaxed},
or working in a continuous space and relying on gradient descent
\citep{guu2015traversing,neelakantan2016neural,yin2016neural,reed2016neural}.
This paper, by virtue of exact dynamic programming, sets the standard.

\paragraph{Acknowledgments.}
We gratefully acknowledge the support of
the Google Natural Language Understanding Focused Program
and the Defense Advanced Research Projects Agency (DARPA)
Deep Exploration and Filtering of Text (DEFT) Program
under Air Force Research Laboratory (AFRL)
contract no.\ FA8750-13-2-0040.
In addition, we would like to thank
anonymous reviewers for their helpful comments.

\paragraph{Reproducibility.}
Code and experiments for this paper are available on the CodaLab platform at
{\small \url{https://worksheets.codalab.org/worksheets/0x47cc64d9c8ba4a878807c7c35bb22a42/}}.

\bibliographystyle{acl2016}
\bibliography{all}

\end{document}